\documentclass[letterpaper,twoside,11pt]{article}

\usepackage{graphicx}
\usepackage{amsmath}
\usepackage{amssymb}
\usepackage{url}
\usepackage{verbatim}
\usepackage{listings}
\usepackage{subfigure}
\usepackage{multicol}

\usepackage{cureport}
\TechReportYear{2011}
\TechReportMonth{September}
\TechReportNumber{1087-11}

\begin{document}

\title{Strange Beta: An Assistance System for Indoor Rock Climbing\\
Route Setting Using Chaotic Variations and Machine Learning}
\author{Caleb Phillips, Lee Becker, and Elizabeth Bradley}
\date{\today}

\maketitle

\begin{abstract}
This paper applies machine learning and the mathematics of chaos to the task of designing indoor rock-climbing
routes.  Chaotic variation has been used to great advantage on music
and dance, but the challenges here are quite different, beginning with
the representation.  We present a formalized system for transcribing
rock climbing problems, then describe a variation generator that is
designed to support human route-setters in designing new and
interesting climbing problems.  This variation generator, termed {\sc
  Strange Beta}, combines chaos and machine learning, using the former
to introduce novelty and the latter to smooth transitions in a manner
that is consistent with the style of the climbs\footnote{This name
  stems from the fact that our system makes use of \emph{strange}
  attractors in order to generate variations on climbing route
  information, which is colloquially called \emph{beta} by climbers.}.
This entails parsing the domain-specific natural language that rock
climbers use to describe routes and movement and then learning the
patterns in the results.  We validated this approach with a pilot
study in a small university rock climbing gym, followed by a large
blinded study in a commercial climbing gym, in cooperation with
experienced climbers and expert route setters.  The results show that
{\sc Strange Beta} can help a human setter produce routes that are at
least as good as, and in some cases better than, those produced in the
traditional manner.
\end{abstract}

\section{Introduction}

Computer assistance in creative tasks, generally the domain of
cognitive science or artificial intelligence research, is a
well-established idea that has attained some success over the past
decades.  For instance, pseudo-random sequences have been used to
create music and art \cite{RandomArt,Hartley1995,Shearer1992,Briggs1992}.  
In this paper, we are concerned with the more-modest goal of assisting humans in a creative task: in particular, using
machine learning and the mathematics of chaos to generate variations on indoor rock
climbing routes.  A similar approach has been
successfully used for generating interesting variations in domains
such as dance choreography and music composition \cite{bradley98a,dabby}.
In these applications, the hallmark ``sensitive dependence on initial
conditions'' of chaotic attractors is exploited to generate a
variation that deviates sufficiently from the input to be unique and
interesting, while at the same time maintaining its basic style.  In
this work, we adapt these techniques to the domain of indoor climbing
routes and validate our approach via a large study in a commercial
climbing gym. We show that computer-aided route setting can
  produce routes that climbers prefer to those set traditionally.

\smallskip

{\noindent The key contributions of this work are as follows:}

\begin{itemize}
\item A language for the representation of climbing problems
\item A parsing framework for mapping the domain-specific natural
  language used by climbers to a succinct set of semantic symbols
in that representation.
\item A machine-learning approach to sequence generation that utilizes
  Variable Order Markov Models (VOMMs).
\item A chaotic variation generator that is designed specifically for
  climbing problems.
\item A user interface that allows human route setters to easily
  explore the space of possible variations and automatically generate
  easy-to-use route plans.
\item Validation of the chaotic variations in a commercial climbing
  gym, using a robust research instrument, in cooperation
  with expert setters and experienced climbers.
\item A publicly available implementation at \url{http://strangebeta.com}
\end{itemize}

The next section of this paper provides a discussion of the most-relevant related
work.  Section \ref{sec:climbing} introduces the problem domain and
defines useful climbing-related concepts; section \ref{sec:overview}
gives an overview of how {\sc Strange Beta} is used.  Section
\ref{sec:language} presents our language for describing climbing
routes and discusses its strengths and limitations; section
\ref{sec:implementation} reviews the mathematics of chaotic variation
and our specific implementation of that strategy.  Section
\ref{sec:ic} describes the {\sc Strange Beta} tool and its results.
Sections \ref{sec:pilot} and \ref{sec:experiment} describe the design,
implementation, and analysis of our two evaluation studies, the first
in the University of Colorado Outdoor Program's climbing gym
and the second at the Boulder Rock Club commercial climbing gym.  In section \ref{sec:learning}, we describe a
parsing strategy for the natural language
that rock climbers use to describe their movements.  We then show how
to train a VOMM on a corpus of climbs and demonstrate how the
resulting model can be used to smooth awkward transitions that may be present in generated routes.  In section
\ref{sec:fin}, we discuss future directions and conclusions.

\section{Related Work}
\label{sec:related}

As mentioned above, the use of computers in creative tasks---for
independent generation and/or in assistance roles---is not a new idea.
Computation has a particularly rich history in music composition, in
both generative and assistant roles \cite{Hartley1995}.  Route setting
for indoor climbing is viewed by its practitioners as a creative task
on par with music composition, requiring substantial expertise in
order to create routes that are both of an appropriate difficulty and
interesting to climb.  A climbing route is a prescribed sequence of
dynamic movements: a sequence of symbols from a complex language not
unlike a dance or a tonal music composition, both of which can also be
viewed as symbol sequences.  All three of these domains---climbing,
music, and dance---have strong notions of ``style,'' but that notion
is very hard to formalize, even for experienced practitioners.  

The challenge in creating a variation on a sequence in any of these domains is to introduce novelty while
maintaining stylistic consonance.  The structure of a chaotic
attractor can be exploited to accomplish this.  Two projects that
apply this idea to the realms of music composition and dance
choreography are especially relevant to {\sc Strange Beta}.  In
\cite{dabby}, Diana Dabby proposed the idea of exploiting the
properties of chaos---the counterintuitive combination of the fixed
structure of a chaotic attractor and the sensitivity of its
trajectories to small changes---to generate variations on musical
pieces.  In her work, a musical piece is codified as a sequence of $n$
pitches (symbols).  A ``reference trajectory'' of length $n$ is
generated from the Lorenz equations, starting at some initial
condition (often $(1,1,1)$, which is not actually on the attractor), and
successive points on the trajectory are assigned to successive pitches
in the musical piece.  Next, a second trajectory is generated with a
different initial condition---say, $(0.999,1,1)$.  Dabby's variation
generator steps through this new trajectory from start to end.  It
examines the $x$ coordinate of the three-dimensional state-space
vector at each point, and then finds the point in the reference
trajectory whose $x$ value is closest to but not greater than that
value.  The pitch assigned to this point is then played.  Variations
generated in this fashion are different from the original piece and
yet reminiscent of its style.  This technique is fairly
straightforward, but the selection of good initial conditions can be
quite a challenge, and that choice strongly affects the results.

{\sc Chaographer} uses
similar ideas to create variations on movement sequences, but with
slightly different implementation of the mathematics and some
necessary domain-specific changes \cite{bradley98a}.  In {\sc
  Chaographer}, a symbol describes the state of 23 joints, which
combine to articulate a body position.  The nearest-neighbor
calculation is generalized to the full dimension of the state space---without the directional restriction in Dabby's work---and care is taken that the
initial condition falls on or near the attractor, which removes some
of choice issues and their implications.  The resulting movement
sequence variations are essentially shuffled concatenations of
subsequences of the original; the stylistic consonance derives from
the subsequence structure, while the novelty derives from the chunking
and shuffling.  {\sc Chaographer}'s companion tool, {\sc MotionMind},
uses simple machine-learning strategies to smooth the potential
dissonance that can occur at the subsequence boundaries
\cite{bradley98c}.  {\sc MotionMind} uses transition graphs and
Bayesian networks to capture the patterns in a corpus of human
movement, then uses those data structures to find a series of
movements that create stylistically consonant interpolations.  In
a simple Turing Test, the chaotic variations were found to be only
marginally less aesthetically appealing to human judges than those
created by human choreographers \cite{bradley}.


\section{About Indoor Climbing}
\label{sec:climbing}

In this section, we give a brief overview of the mechanics of
indoor climbing and climbing route setting.  Appendix \ref{sec:glossary} 
extends this discussion with a glossary of terms. Additional information
can be also be found in \cite{Anderson2004}, which provides a detailed
and practical guide to professional route setting.

While once just for training, indoor climbing has become a popular
sport of its own, with at least one and sometimes several dedicated
climbing gyms in most major cities.  A survey conducted by Roper
Research for the Recreation Roundtable reported that in 2003,
approximately 3\% of the US population\footnote{According to US Census
  data, the US Population was 290,210,914 in July, 2003.}, or 8.7
million people, participated in some sort of rock climbing
\cite{ORCA2003}. 


Indoor climbing walls are designed to mimic rock formations.  They are
often textured, and are covered with embedded ``t-nuts'' so that hand
holds or foot pieces (``jibs'') can be bolted to the surface in
different configurations and orientations.  T-nuts can be arranged on
a geometric grid or in some approximation of a uniform random
distribution.  Holds---generally polyurethane, but sometimes made of
wood, rock, or other materials---come in all shapes and sizes.

Climbers have an informal but fairly consistent language for
describing holds, which involves a relatively small vocabularly of
colloquial terms.  The majority of handholds can be classified into
large open upward-facing pockets (``jugs''), small edges (``crimps''),
or convex rounded holds (``slopers''). There are also more esoteric
shapes (e.g., ``side-pulls'' and ``Gastons'') and composite shapes.
Despite the large number of possible shapes, climbers describe holds
using a readily parseable domain-specific grammar---the topic of
Section \ref{sec:nlp}---that focuses on the holds' function, quality,
and orientation.  

Holds are placed on the wall by experienced route setters to form a
``problem'' or a ``route''---a series of holds with designated start
and ending holds.  Between those endpoints, order is unspecified; part
of the challenge for the climber is to find the right sequence of
holds, which may not be at all obvious.  Climbers use the word
``beta'' to refer to information about how to climb a given route.
Routes differ in length; short ones that do not require a rope for
protection are called ``bouldering'' problems.  Longer routes that
require mostly side-to-side movements are called ``traverses.''  Since
multiple problems coexist on a single wall---and can even share
holds---route setters use colored tape to show which hold is part of
which problem.

Difficulty is determined using a subjective scale. There are several competing
scales in use; in this paper, we employ the widely used American 
scale called the Yosemite Decimal System (YDS). YDS is a subjective 
consensus-based scale, where the easiest problems requiring a rope
are given 5.0 and there is no upper bound, with the currently ``most
difficult'' climb rated 5.15. A postfixed minus or plus (e.g, 5.12-)
indicates that the route is on the ``easier end'' or ``harder end'' of the grade. 
The more-common convention is to use the letters a, b, c, and d 
(where a is easier and d is harder) to more precisely grade a route 
(e.g., 5.10b).

\section{Strange Beta: Overview}
\label{sec:overview}

The rest of this paper describes the details of the design,
implementation, and testing of {\sc Strange Beta}.  By way of context
for that discussion, this section presents a prototypical scenario of
its use by an experienced route setter.  Such a setter might want {\sc
  Strange Beta}'s assistance for a host of reasons, foremost among
which are creativity block (or simply looking for additional
inspiration).  We also imagine that such a tool could also assist in the
training of novice setters.

The first step is to transcribe one or more routes using the
computer-readable language described in section~\ref{sec:language}.  In
doing this, the
route setter can make use of routes from any domain (i.e., outdoors,
indoors, bouldering, etc.)  These routes, which will serve as input to
the variation generator, are stored by the software in a route
database.  Routes transcribed by others can be used as well, but this
is not without problems, as we discuss below.  Variations generated by
{\sc Strange Beta} can themselves be used to generate other
variations, or mixed with additional routes to inject other styles.

When a route-setter is ready to create a new problem, she chooses one
or more routes from the database.  In the simplest scenario, she picks
a single route, but we have found that it is often more interesting to
pick two or more routes to ``mix.''  If the chosen routes are of a
consistent grade and style, then the generated variation will be of a
similar style and grade.  Combining vastly different routes---either
in terms of style or grade---can have unexpected, but often very
interesting, results.  

{\sc Strange Beta} has a variety of controls, 
which set the values of the free parameters in the chaotic 
variation algorithm that it applies to the
chosen routes.  In our implementation, these are presented to the user
in the form of presets (``default'' values and ``more variation''
values).  A setter who is experienced with the software can choose to
vary the initial conditions or parameters of the algorithm in order to
explore alternatives or fine tune the results.

The resulting variation is presented as a ``route plan,'' a sequence
of moves expressed in the language of section~\ref{sec:language}.  To
help the setter make sense of the new route, this route plan
includes a set of annotations that describe how the variation differs
from the original(s).  The setter can print this plan
and use it for direction while setting a route.  During that process,
the setter may choose to make improvisations or corrections to the
variation.  

\section{Route Description Language}
\label{sec:language}

The first challenge in this project was to create a descriptive
language for climbing problems that accurately captures the salient
features of the domain, in sufficient detail to produce interesting
variations, while not being so complex as to form a barrier to use.
{\sc Strange Beta}'s language, which we call CRDL (``climbing route
description language''), is designed to match the epistemology 
of this
domain.  An example CRDL description of a short problem is given in figure
\ref{lst:13}.

\lstdefinelanguage{climb}{morekeywords={L,R,---},sensitive=true}
\lstset{frame=trbl,captionpos=b,float}
\begin{figure}
\begin{lstlisting}{label=lst:13,language=climb,caption=testing 123}
Problem 13 from the CU-hosted RMR CCS
Climbing Competition in March, 2009.
A few large moves between moderate
crimps and slopers with thin/smeared
feet on a vertical wall. Set by Thomas Wong.
Intermediate Difficulty.
- - -
R slopey ledge
L match
R medium crimp sidepull
L diagonal sloper
R crimp (big move)
L sloper (bad) sidewaysish
R crack sidepull
L wide pinch
R match
\end{lstlisting}
\caption{An example CRDL file of a route set by Thomas Wong for a climbing competition at the University of Colorado. 
All the text up until the line containing three hyphens is a header that describes the context of the route for posterity,  
but is ignored (for now) by the variation generator.\label{lst:13}}
\end{figure}

In this formalization, we specifically model the sequence of the hand
movements (L for left and R for right), but leave out the foot
positions. This assumes that a route-setter could easily choose
foothold placements that match the style of the upper-body movements
and produce a route with the desired difficulty.  Similarly, the
wall's characteristics (e.g., steepness) are left out.  As we show in
section \ref{sec:interview}, these assumptions are reasonable; the
steepness is closely associated with difficulty and foot holds can
fairly simply be placed to support desired hand movements.  The
exception to this rule is for ``heel-hook'' and ``toe-hook'' moves,
where the feet are used to hook larger holds.

It is worth noting that these design choices reflect a focus on the
sequence of movements, rather than  on the specific placements of holds on
the wall.  The effect of this is to encourage the person who performs
the transcription of the route to record their subjective
understanding of how the route should be climbed.  This is exactly
what climbers call ``beta.''  This choice regarding the language
design also means that in order to transcribe a route properly, the
person doing the transcription needs to have a good grasp of the
climb, and maybe have climbed or set it themselves.  This, too, is
implicit in climbers' use of the term ``beta.''  CRDL's match to these
understandings and conventions of the domain is intended to make
{\sc Strange Beta} easy for its target audience to use.

We evaluated the success of this language---and indirectly its
accuracy and expressiveness---in two ways: interviews
with users and the consistency of the results.  We found that CRDL is a useful language for climbers and route setters.  As compared
to the work in \cite{bradley,bradley98a}, where individual joint orientations are
modeled explicitly, it is much more free-form and coarser grained.  As
a result, setters found that it is not a chore to transcribe a
climbing problem in CRDL, whereas the notions of specifying individual
joint angles was incomprehensible to human choreographers.  However,
this flexibility comes at the cost of specificity---routes transcribed
with this system might contain a fair amount of
ambiguity\footnote{Students of classical dance notation may notice a
  similarity to Beauchamp-Feuillet notation, which purposely omits
  details under the assumption that a trained dancer would know them
  intuitively.}. In \cite{Anderson2004}, Anderson suggests another language  for
describing routes in climbing competitions. While similar to CRDL, Anderson's maps
also document roughly where holds are placed in space relative to one another.
Because of the complexity of implementing such a system, we have erred on the
side of simplicity and omitted spatial information from CRDL.
One could also impose
more-stringent tests on this language, e.g.: \emph{If a given route A is transcribed by
  one person into CRDL, and that transcription T is used by another
  person to set a second route B, is it true that A is sufficiently
  similar to B that an experienced climber would recognize them as
  being subtle variations on the same premise?}.  This is a matter for
future work.


\section{Generating Chaotic Variations}
\label{sec:implementation}

\begin{figure}
\centering
\includegraphics[angle=-90,width=0.4\columnwidth]{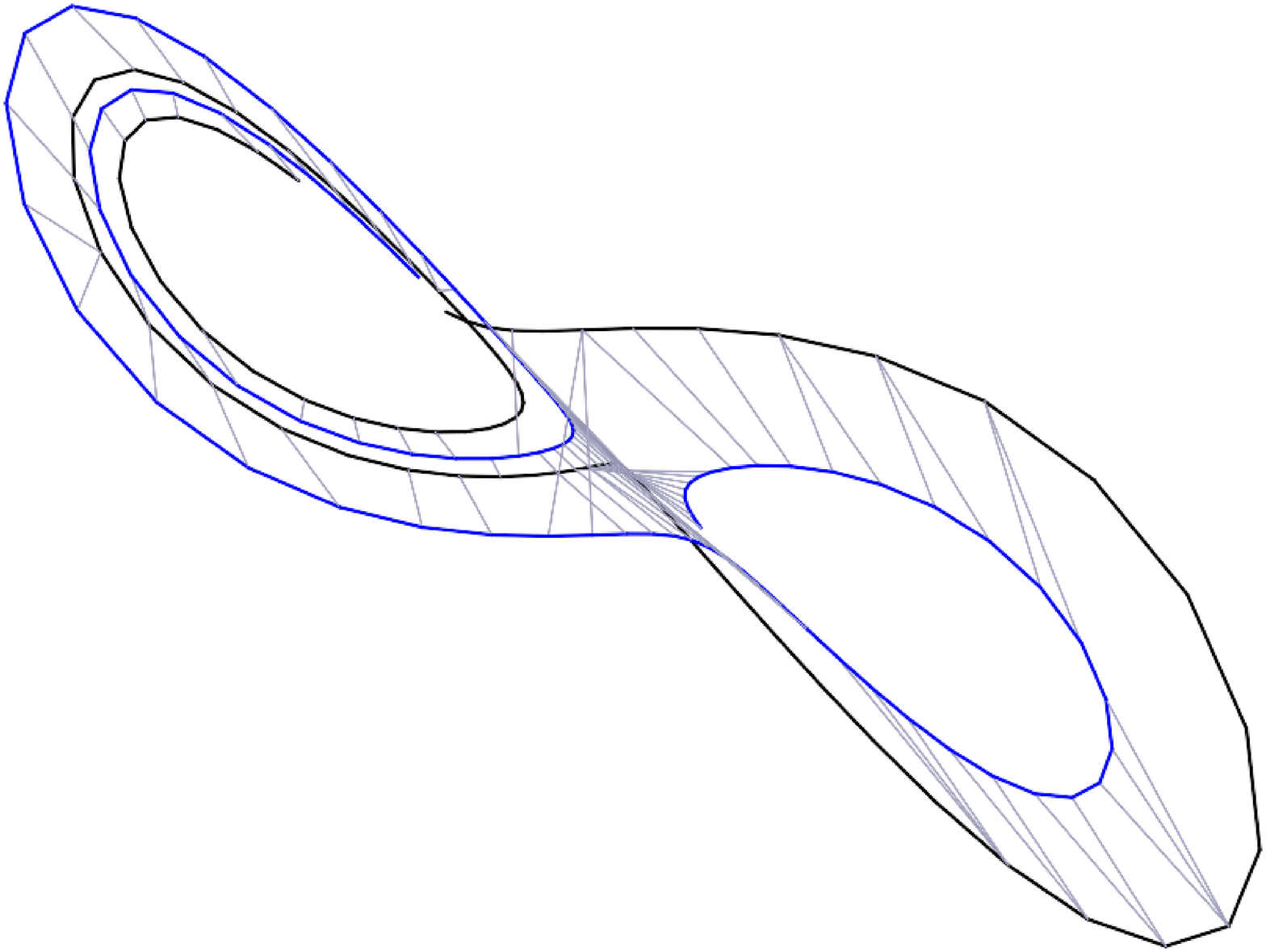}
\caption{Reference (black) and Variation (blue) trajectories for $IC_r
  (-13,-12,52)$ (black), $IC_v = (-16,-13.5,52)$ (light blue)
  projected on the X-Z plane. The gray lines show the associations
  between reference and variation points that produce the
  corresponding variation sequence.}
\label{fig:var}
\end{figure}

To implement {\sc Strange Beta}'s chaotic variation generator, we
follow the same basic design used in \cite{bradley98a} and
\cite{dabby}.  Given a set of ordinary differential equations (ODEs),
some reference initial condition $IC_r$, variation initial condition
$IC_v$, and sequence of input symbols $i = \{i_1,i_2,...,i_n\}$---the
$n$ moves in the ``seed'' route(s)---we use a fourth-order Runge-Kutta
solver to generate two $n$-point trajectories in the state space of
that ODE system: one called $\vec{r}$ beginning at $IC_r$ and one
called $\vec{v}$ beginning at $IC_v$:
\begin{equation}
\vec{r} = \{r_1,r_2,...,r_n\},\;\vec{v} = \{v_1,v_2,...,v_n\}
\end{equation}
We then create the mapping that associates the sequence of input
symbols to the sequence of points in the chaotic reference trajectory
(i.e., $i_1 \rightarrow r_1$ and so on).  Finally, we step through the
variation trajectory point by point, using a Nearest Neighbor
Algorithm (NNA) to find the nearest point in $\vec{r}$ for each $v_k$,
then output the sequence of associated symbols $o =
\{o_1,o_2,...,o_n\}$:
\begin{equation}
o_j = i_k \; s.t. \; k = argmin_l\{d(v_l,r_j)\}
\label{eq:nn}
\end{equation}
\noindent where $d(x,y)$ is some function that calculates the distance
between two points x and y, typically a projected 2-norm (i.e.,
Euclidean distance).  This algorithm is equivalent to the approach
used in \cite{bradley98a}.  In \cite{dabby}, however, the NNA is
unidimensional and directional: it will find the nearest neighbor in
the x-axis if only and only if the neighbor is greater than or equal
to the target.  This causes the algorithm to find no neighbor for some
inputs; it also disturbs the continuity of the variation because its
projection can destroy neighbor relationships.  Dabby
suggests that in this case the user should ``fill in the blanks''\cite{dabby}.
Although we implement both versions of the algorithm, our preference
is for a strict Euclidean NNA of the sort presented in
equation~(\ref{eq:nn}).

The choice of the ODE system and the initial conditions are critical
to {\sc Strange Beta}'s success, as are the parameters of the solver
algorithm.  As in \cite{dabby} and \cite{bradley98a}, we use the Lorenz
system:
\begin{equation}
\label{eq:lorenz}
\begin{split}
x' =& a(y - x)\\
y' =& x(r - z) - y\\
z' =& xy - bz
\end{split}
\end{equation}
\noindent with $a = 16$, $r = 45$, and $b = 4$, arguably the canonical
example of a chaotic system.  We used a solver step size $h=0.015$.
We
chose a reference initial condition near the attractor: $IC_r =
(-13,-12,52)$.  An example trajectory $\vec{r}$ from this initial
condition is shown in figure \ref{fig:var}, together with one of the
variations $\vec{v}$ that we explored in this project.  Dabby investigated other ODE systems, but found the
Lorenz equations to be the ``most desirable''\cite{dabby}.  Similarly, we
considered a R\"{o}ssler attractor, but were unable to convince
ourselves that it generated more-interesting variations---especially
given the short lengths of our trajectories, which are typically on
the order of 30 symbols---so all of the results in this paper use the
Lorenz system.

The choice of symbols is, as alluded to above, another key to the
success of this strategy.  We treated each move---i.e., each line in a
CRDL input sequence like figure \ref{lst:13}---as an individual
symbol.  It is not clear whether it makes more sense to vary the left
and right hands separately or together.  Here, we varied them
together; we will explore the other approach in future work.

\begin{figure}
\centering
\includegraphics[width=0.65\columnwidth]{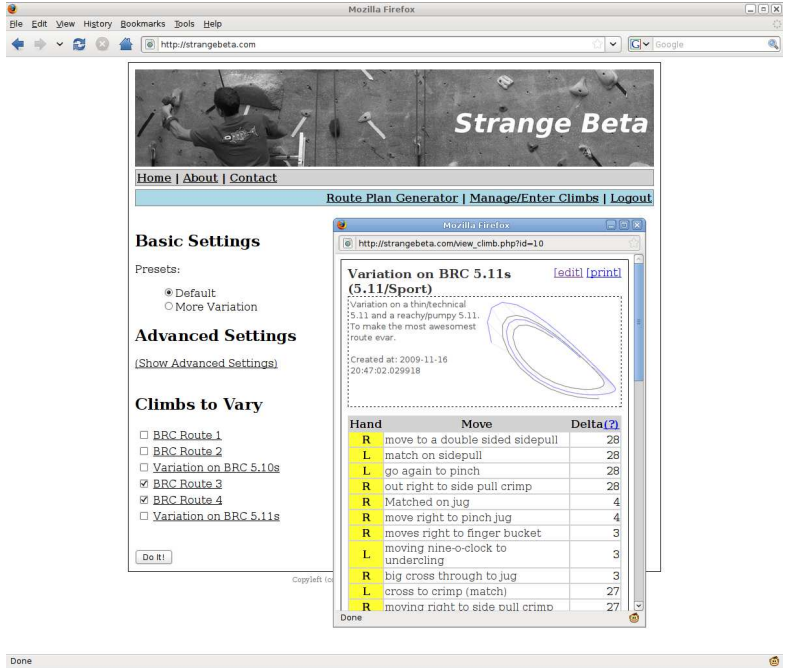}
\caption{Screenshot of {\sc Strange Beta} software being used to
  generate a variation on two input routes.\label{fig:screenshot}}
\end{figure}

{\sc Strange Beta} shares some of the challenges faced by previous
chaotic variation generators.  Route-setters, like musicians or
dancers, are not necessarily familiar with computer-readable formats,
ODE solvers, and chaotic dynamics, so the user interface requires some
real thought.  The software is web based; its output is a ``Chaotic
Route Plan'' that reproduces the input route(s) alongside the
variation.  It specifically indicates which moves in the variation
have been changed and identifies their provenance (i.e., which input
sequence they came from and where).  Figure~\ref{fig:screenshot} shows
a screenshot made during the process of generating a variation on two input
routes.  
The centerpiece of this figure is the chaotic route plan: the varied
sequence together with the annotations about changes and provenance, a
picture of the corresponding trajectories, and some details about how
they were generated.  This route plan can be printed and used by the
route setter, as described in section \ref{sec:overview}.

Another challenge that {\sc Strange Beta} shares with previous
approaches to chaotic variation is novelty.  In both \cite{bradley98a}
and \cite{dabby}, the varied trajectory can only contain a re-ordering
of the specific set of unique symbols in the reference trajectory.
Given the large language of possible climbing movements, this is an
unnatural restriction.  To relax it, we use simple machine-learning
techniques to bring ``new'' or ``unique'' movements into a variation
trajectory, as was done in \cite{bradley98c}.  Details of this
approach are covered in section \ref{sec:smoothing}.

Climbing routes pose some new challenges for chaotic variations as
well:

\begin{enumerate}
\item There are dependencies between some movements. The most obvious
  example is a ``match,'' where a climber places both hands on the
  same hold simultaneously.  How should these dependencies be enforced
  without reducing the chances for interesting variation?
\item Dances and sonatas contain hundreds or thousands of notes and movements, but
  climbing routes are much shorter.  What are the implications of
  this?  How do we generate an interesting variation on a three- or
  five-move problem?
\end{enumerate}

To address the first issue, we simply replace ``match'' moves with
the previous movement of the other hand and add a note to the route
plan: ``(match?)''  This is intended to let the setter know that this
move was used as a match move in the input problem.  We address the
second issue by using multiple climbs as input.  This has the effect
of both increasing the trajectory length and incorporating more
movement types.  When doing this, we generally try to include routes
that are both stylistically similar and of a compatible difficulty.
The result is a variation that takes cues from both routes and is
longer than both. In the scenario where radically different routes are mixed,
the result can be unpredictable. For instance, if a setter chose to combine
an ``easy'' route (e.g., 5.4) with a very difficult route (e.g., 5.13) the resulting
varation is unlikely to be a successful climb, involving sections of intense
difficulty and complexity surrounded by straight-forward movement requiring
little effort (by an experienced climber). Over-long variations are not a
problem; the setter can simply select a chunk of the variation or
eliminate uninteresting sections.

\section{Spelunking for Initial Conditions}
\label{sec:ic}

With an effective chaotic variation generation algorithm in hand, our
next challenge is to help the user choose an $IC_v$ that creates a
variation that is sufficiently different from the input while also
preserving the style.  To this end, {\sc Strange Beta} takes a
brute-force analysis approach.  Given some $IC_r$, we place points on
a $N \times N \times N$ point grid around it, spaced evenly on
intervals of size $s$.  $N$ and $s$ are free parameters of the
algorithm; generally, $N=100$ and $s=0.01$ provide a sufficiently
complex picture of the IC landscape, and so are set as the defaults.
Experienced users can change these values if they wish.

\begin{figure*}[ht]
\centering
\subfigure[Effect]{\includegraphics[width=0.45\columnwidth]{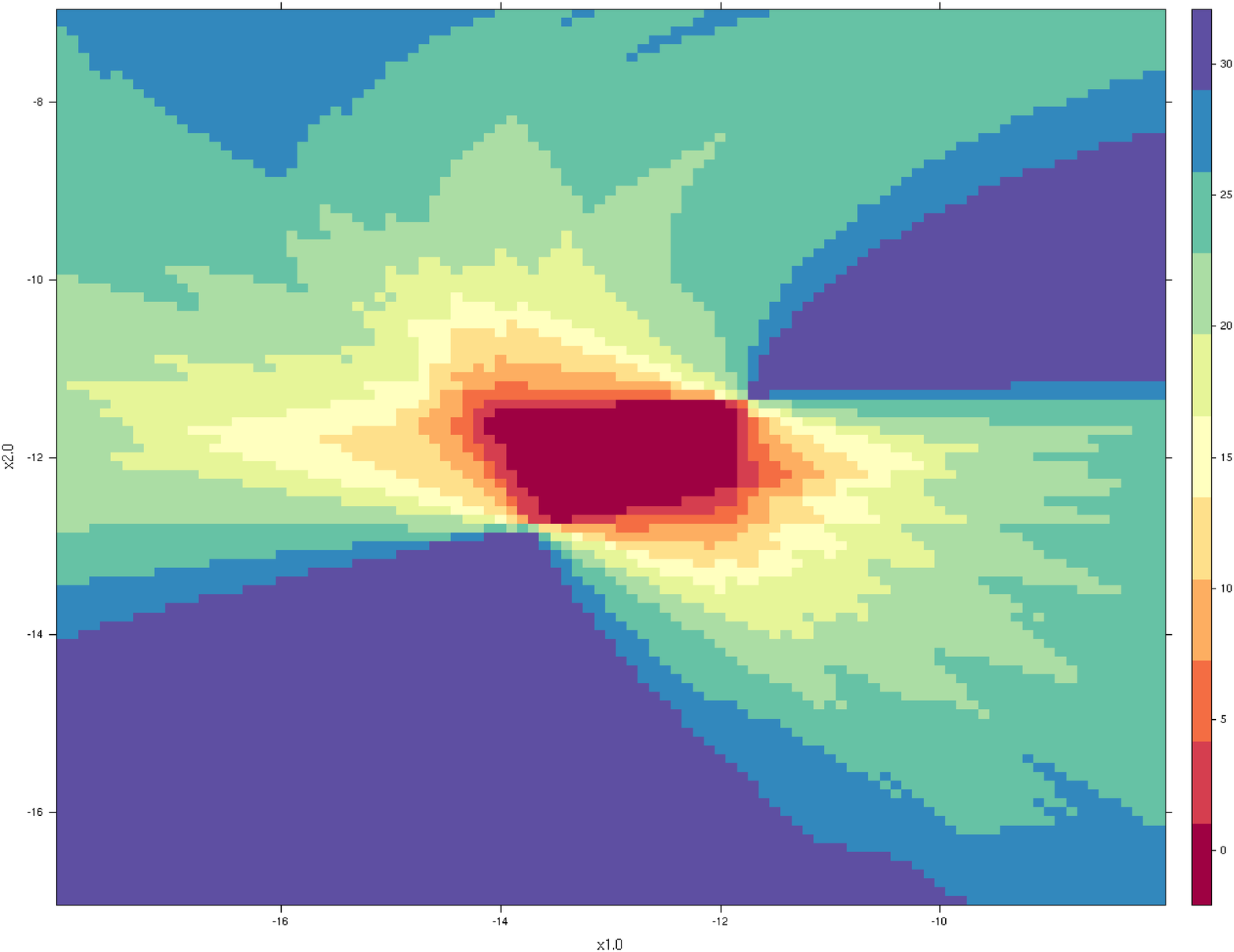}}
\subfigure[Change]{\includegraphics[width=0.45\columnwidth]{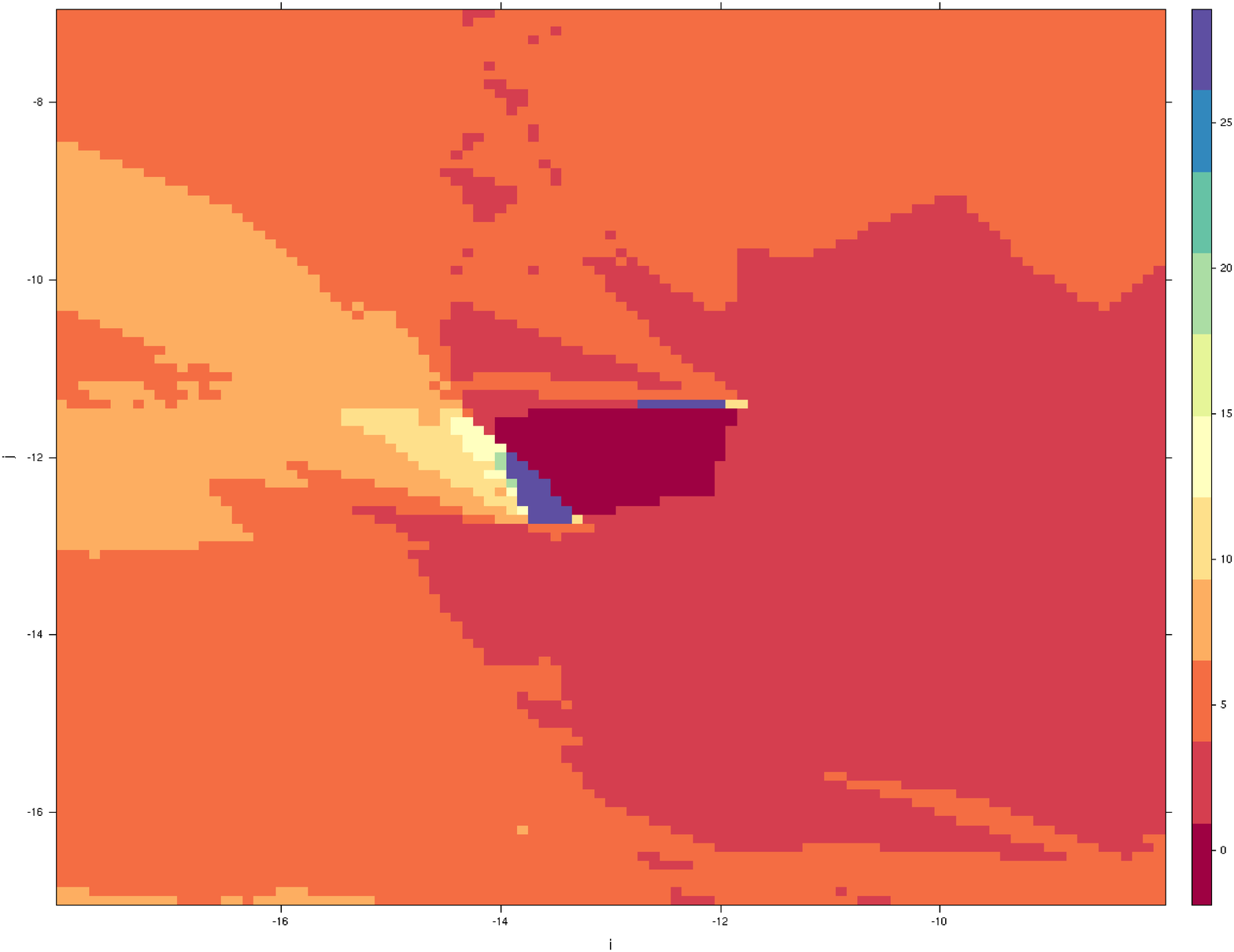}}
\caption{Effect and change for $IC_r = (-13,-12,52)$, $N=100$, $s =
  0.1$ in the x-y plane (i.e., with z held constant at 52) using 2D
  Euclidean distance for nearest-neighbor calculation. In both maps, ``cold'' colors (e.g., blue and purple) indicate
  the largest amount of variation (effect or change) and ``hot'' colors indicate little or no change. Using this color map,
  neutral or average change is colored yellow.}
\label{fig:ec}
\end{figure*}

To help users make sense of this space of possibilities, we color-code
points in that space according to the characteristics of the
variation.  Specifially, we calculate two measures of difference
between the reference trajectory and the variation trajectory:
\emph{effect} and \emph{change}.  Effect is the number of symbols that
would be changed in a chaotic variation.  Change is the average
distance (in terms of index) that those changed symbols would be
moved.
Figure \ref{fig:ec} plots these two metrics for a specific instance.
The effect runs the gamut from no change (the red region) to having
every move changed (the purple region).  However, at those same
points, the change metric tells a different story; instances are
visible where every move is changed, but only by a small amount
(purple effect, red change).  The opposite situation also arises,
where a small number of moves are varied by a large amount.  In
addition to these extremes, there are examples of just about every
condition in between.

Effect and change plots change significantly for different parameter
values.  We generated similar plots for different $s$ values,
different NNAs, and in different projections (clearly we must project
because these metrics are 4-dimensional).  Each combination generates
pictures with different geometry, but the patterns in figure
\ref{fig:ec} are generic and representative, so we chose this $IC_r$
value as the default setting for {\sc Strange Beta}, along with the
$IC_v = (-16,-12,52)$ that is drawn left of center in figure
\ref{fig:ec}, and the 2-dimensional Euclidean NNA calculation.

To help users sift through these results and explore different
$IC_v$s, we implemented an ``IC picker'' tool that preprocesses the
data and finds candidate ICs, given a desired amount of change or
effect.  This automates the otherwise onerous process of varying
parameter values and examining dozens of plots in order to find
appropriate values for one's needs.  This tool also makes it easy to
find multiple---possibly disparate---conditions with similar change
and effect characteristics.  


Generating plots like this requires a great deal of computation and
produces a substantial amount of data---$N=100$ results in 1,000,000
unique ICs and as many runs of an ODE solver on equation (\ref{eq:lorenz}), for instance.
Thankfully, this problem is easily parallelizable.  To compute the results in a tractable amount of time
(30 minutes versus two days), we made use of a 16 node, 128 CPU cluster,
with each of 100 nodes processing 10,000 trajectories. Although this is a tremendous amount
of computation, it only must be performed once for each set of parameters and can
be calculated offline. The resulting static map is then included with {\sc Strange Beta}
so that a user can fine-tune the amount and style of blending by selecting
different ICs from the map.

\section{Pilot Study}
\label{sec:pilot}

With route plans in hand, our next step was to analyze the effectiveness of the
system. After negotiations with the route-setting staff at the University of
Colorado Outdoor Program (CUOP) climbing gym, we were invited to join
them in setting routes. The setters agreed to allow us to set a single route using
the {\sc Strange Beta} software, without interference or assistence from them. 

On 24 April 2009 we set a 29-move traverse called ``Green-13'', which
is based on a variation of two routes: the one in figure \ref{lst:13}
and a longer roped route (a 5.12- at the Rock'n and Jam'n gym in Thornton, Colorado).
These two routes are of similar fairly sustained intermediate
difficultly and similar style, both making use of many crimpers.  We
used {\sc Strange Beta} with $IC_v = (-16,-12,52)$ and $IC_r =
(-13,-12,52)$, then printed out the resulting route plan and brought
it to the gym.  During route setting, we attempted to stay as true to
the variation as possible.  We did deviate from the route plan in a
few places.  For instance, if a move could be modified
slightly\footnote{In this case, by a ``slight'' modification, we mean
  without substantial change to the perceived ``intent'' of the
  variation.} to make use of a hold that was already present on the
wall, we were willing to make that modification instead of cluttering
up the wall with another hold.  We altered the first few moves
of the variation for a different reason: {\sc Strange Beta} placed
the crux of the 5.12- at the beginning of the variation, which is not much fun for climbers.  Finally, the
order of the R and L hands had to be changed several times to make
movements more interesting or avoid awkwardness.

The resulting climb still required a fair amount of human thinking by
the setter, but with a very different flavor and focus than is typical
for route setting.  Typically, a setter approaches the creation of a
new route by trying to think up a particularly interesting move or two
and then building a problem around it.  {\sc Strange Beta}'s route
plan provided us with a useful, interesting skeleton with which to
work.  The staff setters who were present in the gym were clearly
envious.  In \cite{dabby}, Dabby describes her system as an ``idea
generator;'' based on our experience in this pilot study,
{\sc Strange Beta} appears to succeed on that basis.

Assessing results is notoriously difficult in venues where intangibles
like ``style'' play such important roles.  To assess our results, we
placed a questionnaire in the climbing gym.  This questionnaire asked
climbers to compare three routes on each of five metrics, using a
scale from one to five (i.e., a weak Likert scale with a Likert-type
response format). All three routes were of comparable length and
difficulty; one is our Green-13 and the other two were set by expert
human setters.  The participants were not informed about the purpose
of the survey or the difference between the routes.  Six climbers
chose to complete the voluntary questionnaire, one of which was not
blind (one of the staff setters).  Appendix \ref{sec:questionnaire}
provides details about this questionnaire.  Although the sample size
is small, this pilot study provided some useful early feedback.

To analyze the consistency of the instrument, we computed Cronbach's
$\alpha = 0.817$ on the responses.  Doing a per-item consistency
analysis, we determined that answers to question 5 (``Requires
thought/non-obvious beta'') were the least consistent, so we discarded
that question from the analysis, producing a new overall $\alpha =
0.866$.  Although this is a respectable $\alpha$ value, we will not
claim great confidence in this instrument because it is only a
five-item scale---far smaller than is recommended in the literature
\cite{Gleim2003}.  Treating the summed scale
data as ordinal, we can report median attitude values, which are
summarized in table \ref{tab:pilot} for the three routes.  Climb 3---a
traditionally set route---was rated as having the most appropriate,
sustained difficulty, good flow, creativity and interest.  To
determine if the difference in these three medians is significant, we
used a Wilcoxon rank sum test.  The results show that the difference
in medians between {\sc Strange Beta}'s climb and the other two climbs is marginally
significant (p-values of 0.06 and 0.02 for climbs 2 and 3
respectively), but the difference between the two non-chaotic climbs
is not significant (p-value $= 0.37$).  Hence, we can conclude that at
least for this (small) sample, the non-chaotic routes were preferred,
but not by a substantial margin.

\begin{table}
\begin{center}
\begin{tabular}{|c|c|c|}
\hline
Climb & Median Summed Scale Value & Chaotic \\
\hline
1 & 14 & X \\
2 & 15  & \\
3 & 16.5 & \\
\hline
\end{tabular}
\caption{Results of Pilot Survey\label{tab:pilot}}
\end{center}
\end{table}

As is usually the case with pilot studies, these conclusions generated
a host of new questions:

\begin{itemize}
\item Green-13 was set by an inexperienced route setter (author CP).
  How would an experienced route setter employ, and react to, setting
  routes using chaotic variations?
\item What about setting routes at varying levels of difficulty,
  different styles, or different lengths?
\item Is there a qualitative relationship between the choice of initial condition and
  the resulting variation? Can this be controlled, or quantified?
\item Are there other description languages, perhaps more prescriptive,  that
  result in more-interesting variations?
\end{itemize}

To explore some of these questions and to address limitations in
survey apparatus, we performed a much larger study at a commercial
climbing gym.

\section{Experiment and Analysis}
\label{sec:experiment}

Building on the pilot study described in the previous section, we
carried out a second experiment at a large climbing gym, the Boulder
Rock Club (BRC), in collaboration with two expert setters, Tony Yao
(T) and Jonathan Siegrist (J), and the editors of \emph{Climbing}
magazine.

\subsection{Experimental Design and Instrument}

After consultation with these experts and the other setters at the
BRC, we decided to set four routes, two at a grade of 5.10 and two at
a grade of 5.11.  One route of each grade was set using our chaotic
method and the other two were set traditionally.  Using a
questionnaire, we measured the attitude of climbers towards the four
routes.  Again, this study was blind: climbers were not aware of the
research question.  As input to the variation generator, we chose four
existing routes in the gym, two of each grade, both well regarded.
All four routes were transcribed by T.  The two variations were
generated by author CP, using {\sc Strange Beta} with $IC_v =
(-16,-13.5,52)$ and the same $IC_r$ as in the pilot.  We also chose to
skip the first 100 integrated points of the trajectory to further
avoid transient behavior.  These two changes from the pilot study were
intended to produce routes with more variation.

On 30 September 2009, T and J set the four routes using the resulting
chaotic route plans.  Afterward, we interviewed them to record their
thoughts on the experience, which we have summarized below.
Questionnaires were made available at the front desk of the climbing
gym for willing participants and fliers were posted throughout the gym
to advertise the opportunity to participate.  Incentives for
participation were provided by \emph{Climbing} magazine.

Over the course of approximately two weeks, 44 (presumably unique)
climbers completed questionnaires with mean ability, in terms of
typical upper-end outdoor climbing grade, of 5.11c.  Minimum ability
was 5.10; maximum was 5.12d.  On average, participants reported that
they climb indoors between two and three times per week and had been
climbing 12 years, with a minimum of 6 months and maximum of 53 years.
Although we believe this sample to be fairly unbiased and
representative of the population of indoor climbers as a whole, we
cannot claim that this sample is random and hence our analysis is
constrained to making conclusions about the preferences of these 44
participants with regard to the specific four climbs we set.

The questionnaire that we designed to interpret climbers' reactions
and preferences to these routes used standard, well-accepted
techniques for construction of attitude surveys
\cite{Openheim1966,Green2005,Bethlehem2009}.  It was much more
comprehensive than the one used in the pilot study, incorporating
redundancy to enable external consistency checks and explore  our concerns about scale robustness and consistency.
Appendix~\ref{sec:questionnaire} provides additional details about
this questionnaire.  In summary, participants were asked to rate each
climb using a 14-item five-point summative Likert scale as well as a
single direct ranking question.  The five-point response format used
the standard response categories (Strongly Agree, Agree, Neither Agree
Nor Disagree, Disagree, and Strongly Disagree), to which we have
assigned ordinal values of (2,1,0,-1,-2) respectively.  Four of these
questions were negatively keyed so that negative responses indicate
positive attitudes.  These four questions were inverted in
post-processing.  Internal consistency analysis showed that items 1, 9,
10, and 11 produced the greatest inconsistency and were eliminated
from analysis, resulting in a 10-item summative scale with an overall
Cronbach $\alpha = 0.834$ (versus 0.708 before censoring).  This value
indicates that the research instrument is strongly consistent
\cite{Gleim2003,Streiner2003}.

\subsection{Climb Preference}

\begin{table*}
\begin{center}
\begin{tabular}{|c|c|c|c|c|c|c|}
\hline
Climb & Setter & Grade & Med. Summed & Avg. Pos. & Med. Rank & Chaotic \\
      &        &       & Value       & Response \%  & & \\
\hline
1 & J & 5.10 & 6 & 27.44 & 3 & \\
2 & J & 5.10 & 4 & 25.58 & 3 & X \\
3 & T & 5.11 & 9 & 37.23 & 1 & X \\
4 & T & 5.11 & 4 & 26.21 & 3 & \\
\hline
\end{tabular}
\caption{Results of BRC Experiment\label{tab:pref}}
\end{center}
\end{table*}

Interpreting the summed Likert scale data as ordinal, we can compute
the median values for the four climbs, which are given in table
\ref{tab:pref}.  Applying a Wilcoxon rank-sum test to the 5.10 climb's scale
    data, we were unable to reject the null hypothesis that the medians 
    are equal (p-value $= 0.54$).  In the case of the 5.11 climbs, however, we were able
    to reject this null hypothesis: for this sample, the difference
    between medians is significant (p-value $<<$ 0.05).  In other
    words, we can state with confidence that climb 3 is preferred by
    this sample over climb 4, but we cannot make a similar claim about
    the 5.10 climbs, about which the participants were more
    indecisive\footnote{It is worth noting that the setter of the 5.10
      climbs, J, was displeased with his ``chaotic'' route and felt that
      {\sc Strange Beta} had ``helped'' him produce a route that was below the standards he
      typically holds himself to (in fact, he was hesitant to sign his name to it).
      It is interesting to note that, while individual climbers may have preferred one
      route or another, they were generally ambivalent about the two
      5.10 routes.  One possible hypothesis, proposed by the setters
      with whom we worked, is that climbers are less decisive about
      the quality of less difficult routes.}.

Because interpreting summative Likert scale data as ordinal may be
viewed dimly by some conservative statisticians\footnote{Indeed, any
  interpretation of Likert-scale data is a contentious issue
  \cite{Jamieson2004,Carifio2007}.  Although some researchers claim
  that a properly composed and applied summative Likert-scale with a
  sufficient number of questions can produce useful interval-scale
  data, we have erred on the side of statistical conservativism.  To
  this end, we used non-parametric tests and treated the summed scale
  as ordinal, or parametric tests that are robust to skew, in order to
  analyze a continuous variable derived from the ordinal data.}, we
also carried out a similar analysis using a convincingly continuous
variable: percentage of positive ({\sl viz.}, Agree or Strongly Agree)
responses to scale items---an approach common to marketing research,
for example.  Mean values for this variable are in table
\ref{tab:pref}.  A Welch 2-sample t-test on this data produced
congruent conclusions to those described earlier in this paragraph: we
were unable to reject the null hypothesis that the 5.10 climbs have
equal means but we {\sl were} able to reject this null hypothesis with
high confidence in the case of the 5.11 climbs.

As a final indicator of climb preference, we asked participants to
rank-order the four climbs.  The median ranks (where smaller is
better) are listed in table \ref{tab:pref}.  We computed the
inter-grade coefficient of concordance of these ranks using Kendall's
method and found values of $W = 0.01$ with a p-value of 0.59 for the
5.10 climbs and $W = 0.38$ with p-value $<<$ 0.01 for the 5.11 climbs.
These values further serve to indicate that raters are in agreement on
their preference for climb 3 over climb 4, but are not clearly decided
between climbs 1 and 2.

\subsection{Possible Correlating Factors}

In addition to determining climb preference, we also made use of
correlation tests to answer a pair of secondary research questions:

\begin{enumerate}
\item Is preference affected by climb order (i.e., do participants
  rate climbs differently when they are tired)?
\item Is preference affected by climber ability (i.e., do better or
  worse climbers rate climbs differently)?
\end{enumerate}

To address the first question, we used Kendall's $\tau$ on the ordinal
variable and Pearson's r on the continuous variable.  These results
suggested a correlation that is very near zero, both with high
p-values.  From this we concluded that there was no obvious
correlation between climb order and climb preference---or, more
precisely, that we cannot reject the null hypothesis that they are
uncorrelated.  We confirmed this conclusion using the pure ranking
data, which also produced a Kendall coefficient near zero and a large
p-value (near 0.95).  We used the same tests to answer the second
question, producing correlation coefficients (on the order of 0.1)
with p-values greater than $\alpha = 0.05$, indicating that there is
no clear correlation in the data between climber ability and climb
preference.

These two questions correspond to the most obvious sources of data
skew in our specific population.  The $\tau$ and r results indicate that
these concerns are unfounded.

\subsection{Results Summary}

It is clear that the participants of the survey preferred climb 3,
which was set with the assistance of {\sc Strange Beta}, over climb
4, a climb set without it.  In the case of the 5.10 climbs,
participants may have preferred the climb set without the software,
but not by a significant margin.  It is worth noting that the four
climbs that were used as input to the variation generator were
transcribed by T.  From this, one could conjecture that the software
performs best when used by the same setter as did the original
transcription.  Although more work would needed to confirm or deny
this, we suspect that a flexible description format like the one we
have chosen may allow setters to use personal idioms in their
descriptions, preventing portability and reducing the effectiveness
when these same descriptions are used by third parties. \emph{In sum,
  we feel confident in making the claim that when used properly, in a
  scenario where an expert setter feels the use is appropriate, our
  software can assist in producing a route which is at least as well
  regarded as those routes produced without it. And, in some
  cases---indeed, in this study---{\sc Strange Beta} can produce
  routes that are considered superior by climbers to those set purely
  by human experts.}

\subsection{The Setters' Experience}
\label{sec:interview}

The preferences of the climbers are only part of the assessment of
{\sc Strange Beta}; analyzing the opinions of the route-setters is
also crucial to understanding the effectiveness of the system.

To this end, we interviewed the setters after they had finished
setting the four routes.  Although positive about the experience in
general, both setters were hesitant to endorse anything that would
lessen their creative control.  Interestingly, this response is
similar to that of the composers in \cite{dabby}, but at odds with the
more-supportive response dancers had to a similar experiment \cite{bradley}.  One of the two setters, J, found using the generated
route plan to be unwieldy and time-consuming.  This may, again, be a
result of having the other setter, T, transcribe all four input
routes, or it may be a user-interface design issue.  During the course
of the experiment, we identified several ways to improve the route-plan 
format and route-transcription process in order to attend to problems that both
setters encountered using the system.  Both setters, for example,
found some aspects of the output format to be confusing (principally
the inclusion of the input data).  We have since removed this feature from
the output.

To address our initial assumptions about the design of the route
description language, we asked the setters whether it was reasonable
for variations to leave the placement of foot pieces, and the distance
between holds, to their discretion.  They agreed that this was a
reasonable assumption, as usually foot pieces are placed to
accommodate hand movements.  In general, the setters found the
flexibility of the format to be beneficial and allowed them more
creative control overall.

\section{The Grammar of Climbing}
\label{sec:learning}

\begin{figure}
\centering
\includegraphics[width=0.7\columnwidth]{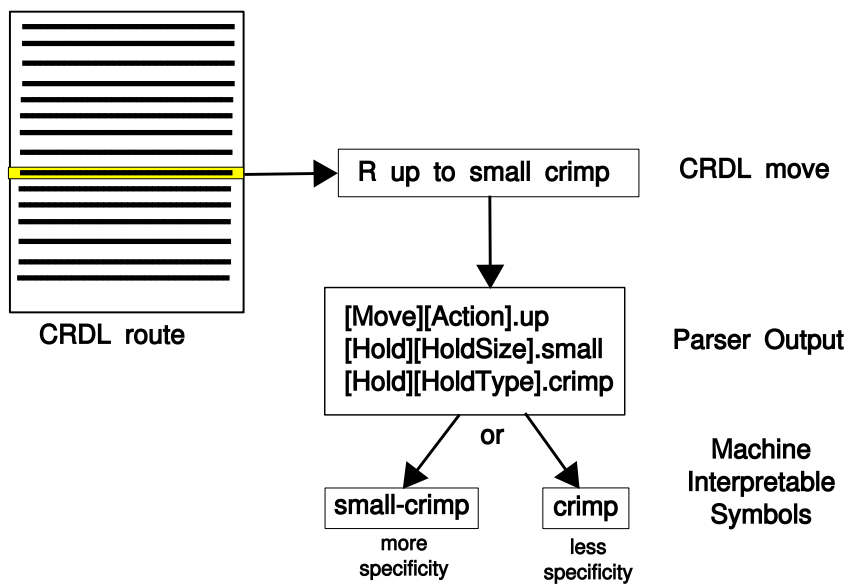}
\caption{Schematic describing the parsing process...}
\label{fig:parsing}
\end{figure}

Movement ``style'' is easy for humans---even non-experts---to
perceive, but it is devilishly hard to formalize.  This issue, which
was explored in the context of dance in \cite{bradley,bradley98c} and
in a variety of good papers in the SIGGRAPH community, also arises
in climbing.  In this section, we present a machine-learning strategy
for capturing the style of a climb.  

The first and most critical step in this process is the definition of
the representation.  The CRDL language in section \ref{sec:language}
is useful for generation of chaotic variations, but it is not rich or
structured enough to support machine learning strategies for this
domain. To address this problem, in this section we seek to parse the free-form
natural language of the CRDL into a set of machine interpretable
symbols. Figure \ref{fig:parsing} shows the strategy taken here: each CRDL
movement line is parsed and then assigned to an unambiguous machine 
interpretable symbol. Section~\ref{sec:nlp} describes climbers' free-form movement
descriptions in a bit more depth and presents a strategy for translating 
those descriptions into well defined parses. Then, section~\ref{sec:setsize}
discusses how parser output can be translated into a set of symbols. By tuning
the number of parsed features that are used, we can control the size and specificity
of the resulting symbol set.

Once a corpus of climbing routes have been translated into symbols, the next
question is how to model them. To this end, section~\ref{sec:smoothing} presents a
variable-order Markov model that we have trained on a substantial corpus of
climbing routes. As an example application of this model, we perform {\sl de novo} 
movement generation (``forward simulation,'' in the parlance of the graphics/animation community)
to interpolate between successive movements, inserting new movements that smooth
awkward transitions that may have been created by {\sc Strange Beta}'s chaotic variation
generator. Because the model is trained on the user-generated corpus of climbing routes, we can
ensure that the resulting interpolation is stylistically consonant with respect to the data
the model is trained on. 

\subsection{A Parser for Climbing-Route Descriptions}
\label{sec:nlp}

After our experiment in the BRC and a subsequent article in
\emph{Climbing} magazine \cite{ClimbingMagazine}, {\sc Strange Beta} gained a fair amount of
exposure in the climbing community.  As a result, a large number of
climbers and route setters entered routes into our public web-based
implementation.  As of 19 April 2011, 250 users had created accounts on
the site, entered 125 routes (comprising 1139 moves), and generated
210 variations (comprising 3800 moves).  This represents a substantial
corpus of human-generated movement sequences that can be used as the
basis of a learning algorithm, but like most natural-language corpora,
it presents some serious challenges to that process.  As an example, consider the
following moves entered by a variety of different users:

\label{page:examples}
\begin{enumerate}
\item ``moving right to a right angle crimp rail''
\item ``bicycle on start and foot chip then match on crimp-jug''
\item ``large horizontal iron cross to a pinch''
\item ``Now move out left 5 feet to huge pinch which looks like a ``F''
\item ``Small Jug''
\item ``cross topostive jug medium to large''
\item ``Dynamic Reach Back Over Sholder About 2 Feet Up, Negitive Side
  Of An I-Beam, Could be Substituted With A Negitive Edge''
\end{enumerate}

At first glance, decomposing these moves into an underlying knowledge representation
seems a very difficult task. The website's unrestricted input provides no way to control the
vocabulary, spelling, or syntax of the entries, yielding data with too
little common structure for simple regular expression matching.  The
variance in word choice and descriptiveness proved to be too ambiguous
for strict approaches like the LL-K algorithms that are used for
parsing programming languages.  Instead we turned to the Natural Language Processing (NLP) literature for techniques robust enough to 
handle the novelty and ambiguity found in descriptions of climbing routes.

Shallow semantic parsing refers to the process of analyzing a text for its basic semantic relationships and propositions.   Current research in shallow semantic parsing centers on learning statistical models for semantic role labeling.  While considered state-of-the-art, these models require large annotated corpora such as PropBank \cite{Palmer:2005fk}, an approximately one-million-word corpus annotated with predicate-argument structures.  By comparison, the relatively small size of our corpus suggests that learning our own semantic parser from data is not feasible.  Existing semantic role labelers are optimized for the newswire text found in PropBank and would be ill-suited for handling the non-standard vocabulary and word usages found in our data.  Furthermore, we are more interested in decomposing user-entered moves into climbing specific semantics rather than the general semantics used by most semantic role labelers.  Consequently, we found more-traditional, rule-based NLP techniques to be better suited to this
corpus of climbing domain language.

For the purposes of parsing route descriptions like the ones in the
list on page~\pageref{page:examples}, we adapted an approach to
natural language understanding that has successfully been employed in
a variety of spoken dialogue systems: semantic frame parsing.  In
frame semantics, an object evokes some scene or related knowledge.  
It is represented as a collection of features
or frame elements, which are in turned filled with values.  In a
travel domain, for example, the \emph{Flight} frame might consist of
frame elements such as \emph{Origin}, \emph{Destination},
\emph{Departure Time}, and \emph{Airline}.  From a sentence, a frame
semantic parser identifies the appropriate frame and extracts the
words that fill the frame elements.  Thus sentences like ``I am flying
from Seattle to Denver on August 20th on Frontier Airlines.''  and
``I'm coming back to Denver from Seattle on the 20th of August via
Frontier'' would parse into an identical set of frame-elements. This approach
is well suited for our task because its robustness in the face of ungrammatical
constructions and the relative ease of authoring grammars makes it well
suited for the domain-specific, fragmented, and variable
descriptions generated by our user base.


Our key NLP task was to parse a description of a sequence of climbing
moves into the appropriate frame semantic representation.  To do this,
we used the Phoenix Parser \cite{Ward1994}, a flexible semantic frame
parser built using recursive transtion networks.  We began by
authoring a context-free grammar for recognizing different classes
of moves, and then we iteratively refined it until the system was able
to parse a large percentage of our user-generated text. At each step,
we run the entire corpus of moves through the parser and inspect those
move descriptions that failed to parse, and make small adjustments to the grammer
to fix what is missing (or wrong). When the grammer is able to parse
more than 95\% of the input corpus, we stop making improvements. By stopping
short of 100\% acceptance, we avoid overfitting the grammar to the input data.

\begin{figure}
\begin{center}
\includegraphics[width=\columnwidth]{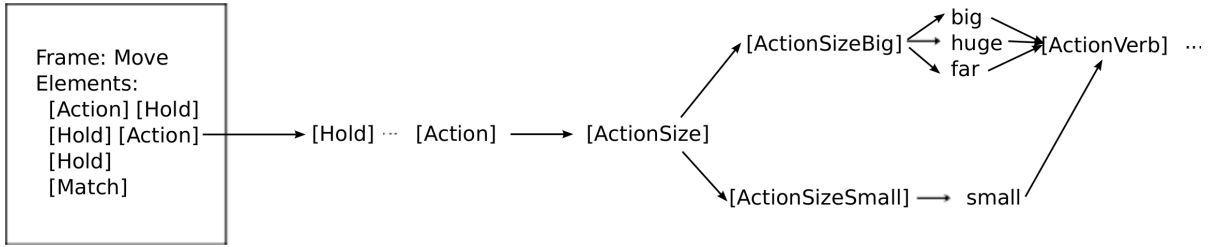}
\caption{Illustration of a small part of the Phoenix
  Grammar\label{fig:grammar}}
\end{center}
\end{figure}

In designing the Phoenix grammar, we chose to use a single \emph{Move} frame, which
consists of \emph{Action}, \emph{Hold}, and \emph{Match} frame elements to
capture the salient features of the route descriptions. Figure \ref{fig:grammar}
shows an illustration of a small part of the resulting grammar and appendix \ref{sec:grammar}
provides the complete grammar. Given the larger goal of creating an assistance system, it is important that the 
frame elements' semantic decompositions reflect differences in climbing difficulty; consequently, 
the grammar rules were written to discriminate moves by hold quality, hold size, and action size.
To illustrate this breakdown consider the phrase ``large horizontal iron cross to a pinch''.  If the grammar is correctly defined, the parser would recognize ``large \dots cross'' as a big action and pinch as a type of hold.
Using this grammar, we are able to successfully parse 95\% of the free-form 
user-entered moves. Without a hand-parsed ``oracle'' corpus, validation must be done manually. To this end, we sanity
check the parser's output, and make any final corrections to the grammar as necessary

As a complete example of how the parser functions, consider the following ``noisy'' CRDL move:
\begin{verbatim}
R moving right to a right angle crimp rail 
\end{verbatim}
\noindent Given this input, our parser generates the following output:
\begin{verbatim}
PARSE_0:
IsMove:[Move].[Action].[ActionVerb].[ActionVerbSmall].[ActionVerbSmallT].right
       [Hold].[HoldShape].[HoldShapeGood].[HoldShapeGoodT].right angle
       [HoldType].[HoldTypeT].crimp
IsMove:[Move].[Hold].[HoldType].[HoldTypeT].rail
END_PARSE
\end{verbatim}
In this example, the parse is actually ambiguous, resulting in two matching
{\tt IsMove} frames: ``right-right-angle-crimp'' and ``rail''.  In our application,
this ambiguity is not a problem since  these two parses clearly relate
to different parts (or features) of the same move. Hence, in post-processing a script combines the two
parses to acheive the hybrid move symbol: ``right-right-angle-crimp-rail'', which
matches what the user was (presumably) attempting to describe\footnote{A 
``crimp rail'' is how a climber might describe a long shallow hold that requires
crimping. In this particular move, the hold appears to be vertical in orientation 
(right angle) and is located to the right.}. Conveniently, our grammar does not 
need to include explicit information about hybrid holds, so long as it includes 
the basic vocabulary of which they can be composed.

\subsection{Tuning Specifity with Symbol Set Size}
\label{sec:setsize}

In the previous section, we discussed how parser output can be used explicitly to generate
machine interpretable symbols. In this section, we will consider using a subset of parsed
features to generate multiple symbol sets at different levels of specificity. For instance,
consider the final example from the previous section of the symbol: ``right-right-angle-crimp-rail''.
In some applications it may be only useful to know the hold type (``crimp-rail'') and not the direction
(``right'') or the positioning of the hold (``right-angle''), while some other applications may
operate on these more specific features as well. By limiting the specificity of the resulting symbol
set, we can provide more flexibility in the resulting model(s) and at the same time reduce
the complexity of the modeling task by limiting the size of the input language. 

\begin{table}
\small
\begin{center}
\begin{tabular}{|c|l|c|l|}
\hline
Symbol Set & Description & N & Example Symbol \\
\hline
1 & Hold Types & 68 & sloper \\
2 & Hold Types with Descriptors & 182 & edge-small-sloping \\
3 & Hold Types with Quality Booleans & 110 & pinch-(big move)-(good hold) \\
4 & Hold Types with Descriptors & 220 & pinch-small-sloping-(cross) \\
  & and Quality Booleans & & \\
\hline
\end{tabular}
\end{center}
\caption{Descriptions of the four symbol sets. $N$ is the number of unique symbols present in our parsed data.}
\label{tab:symbols}
\end{table}

To enable flexibility in modeling, we define four symbol sets of increasing specifity, which are described in table \ref{tab:symbols}. Each
symbol set focuses on a set of dinstinct features that we believe are likely to be most meaningful in this domain, and stem
from the structure of the parser grammar. The first symbol set simply assigns a symbol to each type of hold, without any concern for attributes, position, orientation, or movement.  The second set includes parsed size and shape attributes.
The third and fourth are modifications of the first and the second, respectively, that incorporate a set of three special boolean
attributes that explicitly describe the quality (difficulty) of the move:

\begin{itemize}
\item Is the move a cross?\footnote{Moves that involve a cross require the climber to
  cross one arm over (or under) the other to reach a given hold. This is a
  common movement in difficult climbs and we found it useful to model it explicitly.} 
  This is set to true if that word appears  in the data entry. 
\item Is the hold ``good''?  This is set to true if the sum of hold
  descriptors in the data entry is biased towards positive statements
  (i.e., ``big,'' ``good,'' etc.) and false otherwise.
\item Is the move ``big''?  This is set to true if the movement verb
  in the data entry is modified with a size attribute.  
\end{itemize}
The rationale behind this procedure is that a move that involves a ``bad'' hold,
a ``big'' movement, or a cross, is almost always more difficult than a
move without these features.  


In any problem of this type, representational
accuracy---and thus the results of any algorithms that operate on the
resulting data---depends heavily on the user-friendliness of the
formal language.  Since the natural language used by climbers in
describing their routes is so varied and unstructured, that issue is
of critical importance here.  Judging by the success of this
strategy---successful parsing of more than 90\% of inputs from
hundreds of climbers distributed around the world---our grammar
appears to be doing justice to this domain-specific language.  However,
a fuller evaluation of these issues will require more data, which 
we are in the process of gathering via the website. 



\subsection{Modeling the Grammar of Climbing}
\label{sec:smoothing}

A ``generative'' model trained on a corpus of climbing routes that have been parsed as described in
the previous sections, can be used to generate new climbs---or chunks
of climbs---that have the same style as the others in that corpus. This has
numerous applications, but here we will explore the problem of smoothing
transitions between movement subsequences.

%

To capture the grammar of climbing, we look to Markov models.  This amounts to
viewing the climb-setting process as a state machine in which the
probability of transitioning to the next state (move) is a function of
the current state.  While the state-based nature of this approach
seems appropriate for our domain, we could not convince ourselves that
a {\sl single} move provided sufficient context to make a reasonable
prediction.  In order to consider more context, we used Variable Order
Markov Models (VOMMs), which determine the probability of moving to
successor states from the current state and a variable number of prior
states.  VOMMs have been used widely in compression algorithms (e.g,
\cite{Bell1989}), but only recently in the context of prediction.
\cite{Begleiter2004}, for instance, use VOMMs to learn a sequence and
predict the most likely next state. They study how several VOMM
learning algorithms handle different types of input, finding the
``decomposed'' Context Tree Weighting (DE-CTW) to be a good
general-purpose model.  Significantly, DE-CTW was the best at learning
music scores and predicting the next most likely note sequence---a
task that bears substantial resemblance to what {\sc Strange Beta} does.


Using the implementation from \cite{Begleiter2004}, we trained a VOMM
on the symbol sequences  generated by
the parser described in section \ref{sec:nlp} from the entire corpus
of user-entered data on the {\sc Strange Beta} website.  In order to
study inter-user differences and allow users to request models that
match their individual styles, we also trained a separate VOMM model
for each user's data.  (The full-corpus model, in contrast, captures
some average of the overall style---an amalgam of the styles of all of
the individuals who contributed.)  There are other interesting cross
sections that one might try: a model of climbs set by experts, models
for different climb difficulties, models for different geographic
regions\footnote{One might hypothesize that the style of climbing in a
  given region is affected by the style of climbing necessitated by
  the surrounding geology}, or models of different types of climbing
(sport climbing, top-roping, bouldering or traversing), for instance.

A VOMM trained in this fashion can be used for a variety of
interesting purposes, including recognition, forward simulation, and
interpolation.  Given a sequence of moves, it can provide a estimate
of the likelihood that that climb was produced by that model:
\begin{equation}
L = -log_2(P(sequence|model)
\end{equation}
Forward simulation is a matter of creating a sequence of moves from a
given starting point that minimizes the negative log likelihood value
$L$.  Interpolation adds another constraint: given a starting move, an
ending move, and the surrounding k-sized context of moves, is there
some number of moves we can insert between the two that will minimize
the negative log likelihood of the entire sequence?

We experimented with climbing-route interpolation using the VOMM model
trained on the parsed corpus.  To find an interpolation sequence, we
performed an exhaustive search, looking for the set of
inserted symbols that minimize the negative log likelihood.  This
search involves visiting $N^j$ nodes, where $j$ is the maximum-sized
insertion considered and $N$ is the size of the symbol set (from table
\ref{tab:symbols}).  This search is exponential in $j$, but running it
with $1 <= j <= 2$ is useful for this domain.  (We found that when
setters judge a route to be jerky or disconnected, for
instance---which can arise in {\sc Strange Beta} as a side effect of
the chaotic variation strategy---they generally inserted only one or
two extra holds to ``smooth'' the transition.)  We have yet to perform
a full formal evaluation of these results, but the preliminary results
are promising.  Like a human expert, this VOMM-based scheme inserts
common movements---simple moves to a jug or crimp---to ``fix''
discontinuities.  Whether or not this produces climbs that are
actually more ``pleasant'' or ``natural'' to climbers is an open
question.

\section{Conclusions and Future Work}
\label{sec:fin}

In this paper, we have applied chaotic variations and machine learning
to a new domain with some unique and interesting challenges: indoor climbing route setting.
We have proposed new ways of exploring the space of possible
variations and validated the results in two user studies.  We found
that our chaotic variation strategy is a useful assistant to human
experts, helping them produce routes that are at least as well
regarded as those set traditionally.  We proposed a formal
representation for climbing route description and a parsing strategy
for transforming the informal movement descriptions provided by users
into a set of unambiguous symbols.  We then trained a variable-order
Markov model on a substantial corpus of data gathered from the broader
climbing community and used it to create stylistically consonant
climbing sequences.

Though {\sc Strange Beta}'s chaotic variation facilities have been
fairly well worked out and assessed, the machine-learning work
described in the previous section is only a beginning.  There are
particularly interesting opportunities for future work concerning
representations, methodologies, and incorporation of domain knowledge.
A formal evaluation of the symbol sets of table~\ref{tab:symbols} and
the parser framework of which they are targets could lead to a
more-effective knowledge-representation framework, which would in turn
support better models.  It would also be interesting to compare and
contrast VOMMs to other learning algorithms for this application.
Assessing how well {\sc Strange Beta} works in different environments
and for a large number of different setters will require a much more
substantial experimental evaluation than we have done here.

Finally---and of most interest to us---is the incorporation of domain
knowledge into {\sc Strange Beta}'s mathematics and models.
Identification of the crux, the most difficult section of a route, is
of particular importance.  We know from our surveys that the quality
of a crux is important to a climber's impression of the route.  If we
could identify the crux of a route, then we could determine whether or
not a chaotic variation or a VOMM-generated route has one---and if it
is of a reasonable size, shape, and position in the overall route.
Ultimately, we would like to explicitly address the question of {\sl
  how} human setters create interesting short sequences and use that
understanding as a basis for a machine-learning solution.  Existing
research on biomechanical models for equilibrium acquisition while
climbing \cite{Quaine1999} could support this endeavor, as it offers
explicit models for climbing-related movement in route generation.
These explicit mechanical models might be expanded with cognitive
models for how climbers visualize climbs---a combination of not just
movements, but also specific application of force and effort
\cite{Smyth1998}.

Gathering the data to support all of these research threads will
require the continued interest and participation of the climbing
community.  To this end, we continue to prototype new features on the
publicly facing implementation of the system at
\url{strangebeta.com}.  These data, and the associated results,
would also contribute to existing academic research on rock climbing.
Research on the exercise physiology of difficult climbing, for
instance, has produced well-defined training guidelines for climbers
\cite{Watts2004}.  {\sc Strange Beta} could dovetail nicely with
this, generating route variations aimed at specific training goals.

Overall, we believe that chaotic variations provide great promise in
the realm of creative processes.  Though there are many open questions
and much to be done, the work described here serves two important
purposes from the standpoint of the climbing community.  Firstly, it
is a large step forward in terms of creating a functional prototype of
such a system.  And secondly, and perhaps most importantly, it has
convinced us and others that chaotic variations are a useful technique
in this domain.  We are uncertain whether our approach to route
setting will be widely adopted, in large part because expert setters
enjoy the creative challenges of setting unique and interesting
problems from scratch.  However, we see promising applications when
creativity block strikes or when teaching novice setters.

\section*{Acknowledgements}

We would like to thank the climbing gyms and route setters who
supported this work for their help---particularly setters Tony Yao and
Jonathan Siegrist at the BRC and Hana Dansky and Thomas Wong at the
CUOP climbing wall.  Their thoughtful feedback, and their help in
setting routes, allowing us to access their facilities and solicit
their patrons was essential to this project.  Matt Samet, the Editor
in Chief of \emph{Climbing Magazine} was very helpful in the
organization and design of the BRC experiment; without his enthusiasm
for the project, we would not have been able to obtain the results we
have presented here.  Finally, Dr. Jeffrey Luftig provided crucial
criticism and suggestions regarding the design of our experimental
instrument and the subsequent statistical analysis, and Dan Knights
provided early and useful insight on the possibility of applying
machine learning to rock climbing sequences.

\appendix

\section{Glossary of Climbing Terms}
\label{sec:glossary}

\begin{itemize}
\item \textbf{Beta} - Information about how a route/problem must (or
  can) be climbed.
\item \textbf{Crimper} - A shallow hold that may only support the tips
  of the fingers and hence might need to be ``locked off'' (where the
  thumb reinforces the position by pressing down on the forefingers)
\item \textbf{Crux} - The most difficult move(s) of a route.
\item \textbf{Gaston} - The opposite of a sidepull, with the gripping
  surface facing inward ({\sl viz.,} the grip used to pry open an
  elevator door).
\item \textbf{Jib} - A very small knob-like hold, usually used as a
  foot piece.
\item \textbf{Jug} - A fairly deep hold whose geometry is similar to
  that of a steep-walled pot or jug.
\item \textbf{Match} - A hold that is held with both hands
  simultaneously.
\item \textbf{Problem} - Another term for a climbing route which more
  directly captures the often puzzling nature of climbing routes.
\item \textbf{Redpoint} - To climb from start to finish without
  falling, typically placing protection on the way.
\item \textbf{Sidepull} - When a hold is positioned so that its main
  grip ping surface is away from the climbers body ({\sl viz.,} the
  grip one would use to close a sliding glass door).
\item \textbf{Sloper} - A more-rounded hold that is gripped with the
  palm of the hand or pads of the fingers, to create friction.
\end{itemize}

\section{Questionnaire Design and Consistency}
\label{sec:questionnaire}

The questionnaire used in the pilot study at the University of
Colorado climbing gym used a five-question Likert scale with a
five-item Likert-type response format to determine the attitude of
climbers regarding each climb.  The questions are listed in table
\ref{tab:likertpilot} along with Cronbach's $\alpha$, a measure of
internal consistency of the responses.  
\begin{table*}[t]
\begin{center}
\begin{tabular}{|c|l|c|c|c|c|}
\hline
No. & Question & $\alpha$\\
\hline
1 & Appropriate Difficulty & 0.725 \\
2 & Sustained Difficulty & 0.727 \\
3 & Has Good Flow/Seems Consonant & 0.810 \\
4 & Is Creative/Has Interesting Moves & 0.755 \\
5 & Requires thought/Non-obvious Beta & 0.866 \\
\hline
\end{tabular}
\caption{The pilot study instrument: a five-item Likert scale intended
  to assess participants' attitude towards a given climb.  Cronbach's
  $\alpha$ is a measure of internal
  consistency.\label{tab:likertpilot}}
\end{center}
\end{table*}
The $\alpha$ reported in this table is the {\sl overall} Cronbach's
$\alpha$, with the given question removed---hence a number larger than
the overall $\alpha = 0.817$ (the $\alpha$ obtained for all questions, without any censored) 
indicates that censoring this question would improve the scale's consistency \cite{Gleim2003,Streiner2003}.
We also asked participants to state whether they were a Beginning,
Intermediate, or Advanced climber, but did not use this information in
our analysis.

The questionnaire used in the larger BRC study used a 14-question
Likert scale with a five-item Likert response format. As compared to the
5-question survey used in the pilot study, this survey includes many more questions
which are separated into the same basic categories tested in the survey. The reasoning
here is that a larger number of questions allows for greater redundancy and ability
to measure internal consistency. In addition, we also introduced a rank-ordering
question to serve as an external consistency metric, to compare the the Likert-scale
results, and expanded the demographic questions to allow for further investigation
of possible correlating factors. Finally, we asked participants which order they climbed
the routes in, since we hypothesized that the larger number of experimental routes (4 versus 1)
might lead to ordering effects. Table \ref{tab:likert} lists the questions along with their internal
consistency metrics.

\begin{table*}[t]
\begin{center}
\begin{tabular}{|c|l|c|c|c|c|}
\hline
No. & Question & $\tau$ & $\alpha$ & Neg.\\
\hline
1 & Too easy for the grade & -0.108 & 0.750 & X \\
2 & Too difficult for the grade & 0.184 & 0.703 & X \\
3 & Difficulty is consistent throughout the climb & 0.220 & 0.696 & \\
4 & Requires thoughtful/nontrivial beta & 0.117 & 0.707 & \\
5 & Has good flow throughout & 0.536 & 0.661 & \\
6 & Appears to be well thought out & 0.556 & 0.650 & \\
7 & Is creative/has interesting moves & 0.480 & 0.657 \\
8 & Climbs awkwardly & 0.413 & 0.668 & X \\
9 & Good variety of handholds/types of grips & 0.161 & 0.705 & \\
10 & Has a definite crux & -0.121 & 0.753 &  \\
11 & Crux is technically engaging & 0.043 & 0.719 & \\
12 & Has an unpleasant/stopper crux & 0.173 & 0.701 & X \\
13 & Is a route I would climb again & 0.560 & 0.645 & \\
14 & Is a route I would recommend to others & 0.633 & 0.638 & \\ \hline
\end{tabular}
\caption{The BRC study instrument: a 14-item Likert scale intended to
  assess participants' attitude towards a given climb.  Cronbach's
  $\alpha$ is a measure of internal consistency; $\tau$ is a measure of
  correlation, which is being used a secondary measure of consistency,
  and questions that are negatively keyed (i.e., a positive response indicates
  a negative attitude) are flagged.\label{tab:likert}}
\end{center}
\end{table*}
The $\alpha$ value, again, should be considered relative to the
overall $\alpha$ of 0.708.  $\tau$ is the Kendall's $\tau$ correlation
coefficient for each question's rating, as correlated with the overall
rating. Hence, a question with a large $\tau$ and large $\alpha$ relative to the mean,
are generally consistent with the overall results \cite{Openheim1966}. Some questions are negatively keyed
to avoid bias from having all questions be repetitively positive or negative. For instance,
question 1 asks whether the route is too easy for the grade (a negative statement) and
question 5 asks whether the route has good flor (a positive statement).  In addition to the
Likert scale, this questionnaire requested some domain-specific
demographic information and asked participants to rank-order the
climbs (which served as an external consistency check) and list the
order in which they climbed them (which served to expose any ordering
bias).  These questions are listed in table \ref{tab:oq}.
\begin{table}[h]
\begin{center}
\begin{tabular}{|l|}
\hline
Question \\
\hline
Years climbing? \\
Years climbing in a gym? \\
Hardest indoor redpoint? \\
Hardest outdoor redpoint? \\
Days per week climbing outside? \\  
Days per week climbing at the BRC? \\
Typical indoor grade range? \\
Typical outdoor grade range? \\
In what order did you climb the routes (e.g., 1,4,3,2) ? \\
What is your overall ranking of the routes from \\
~~~best to worst (e.g., 4,2,1,3)? \\
What is your favorite ice cream flavor? \\
\hline
\end{tabular}
\caption{Demographics and other questions from the BRC
  study. Climber ability, experience, and order were used to asses 
  possible correlations with climb preference. Climb ranking
  was used as an external consistency metric. To the disasppointment of your
  authors, no significant correlation was found between ice cream preference 
  and climber ability.\label{tab:oq}}
\end{center}
\end{table}

\clearpage

\section{Phoenix Grammar Specification for Climbing Routes}
\label{sec:grammar}  


\begin{multicols}{2}

\begin{verbatim}
[Move]
    ([Action] [Hold])
    ([Hold] [Action])
    ([Hold])
    ([Match])
;

[Match]
    (match)
;

[Hold]
    ([HoldSize] [HoldShape] [HoldType])
    ([HoldShape] [HoldSize] [HoldType])
    ([HoldShape] [HoldType])
    ([HoldSize] [HoldType])
    ([HoldType])
;

[HoldSize]
    ([HoldSizeBig])
    ([HoldSizeSmall])
;

[HoldSizeBig]
    ([HoldSizeBigT])
    ([Not] [HoldSizeBig])
;

[HoldSizeBigT]
    (big)
    (good)
    (manageable)
    (managable)
    (deep)
    (positive)
    (goodish)
    (okay)
    (ok)
    (solid)
    (decent)
;

[HoldSizeSmall]
    ([HoldSizeSmallT])
    ([Not] [HoldSizeBig])
;

[HoldSizeSmallT]
    (mini)
    (shallow)
    (small)
    (bad)
    (razor)
    (shitty)
    (tiny)
    (micro-dick)
    (transition)
    (nonexistent)
;

[HoldShape]
   ([HoldShapeGood])
   ([HoldShapeBad])
;
   =20
[HoldShapeGood]
    ([HoldShapeGoodT])
    ([Not] [HoldShapeBad])
;

[HoldShapeGoodT]
    (starting)
    (vertical)
    (bulbous)
    (angle)
    (sideways)
    (double sided)
    (right angle)
    (left angle)
;

[Not]
    (not)
    (no)
;

[HoldShapeBad]
    (roof)
    (slopey)
    (sloping)
    (vertical)
    (finger)
    (diagonal)
    (angled)
    (gaston)
    (flat)
    (downward)
    (down ward)
    (open hand)
    (openhand)
    (reachy)
;

[HoldType]
    ([HoldTypeT])
    ([UnderCling])
    ([SidePull])
    ([FootHook])
    ([GenericHold])
    ([Layback])
    ([Mantle])
    ([Jib])
;

[HoldTypeT]
    (jug)
    (pocket)
    (crimp)
    (edge)
    (sloper)
    (cobble)
    (crimper)
    (crimpbeam)
    (beam)
    (layback)
    (horn)
    (ball)
    (boobies)
    (slope)
    (pinch)
    (bucket)
    (rail)
    (ear)
    (cup)
    (flake)
    (thumbcatch)
    (slot)
    (gaston)
    (dish)
    (ledge)
    (incut)
    (teeth)
    (arete)
    (tufa)
    (hand jam)
    (fist jam)
    (finger jam)
    (mono)
    (offwidth)
    (chicken head)
    (knob)
    (handle)
;

[Mantle]
   (topout)
   (top out)
   (mantle)
   (finishing hold)
   (top)
   (finish)
;

[GenericHold]
   (hold)
   (hand)
   (feature)
   (grip)
   (start)
;

[Layback]
   (lay back)
   (layback)
   (lie back)
   (lieback)
;

[Jib]
   (jib)
   (gib)
   (churd)
;

[SidePull]
    (sidepull)
    (side pull)
;

[UnderCling]
    (undercling)
    (under cling)
;

[FootHook]
    (heel hook)
    (heelhook)
    (toe hook)
    (toehook)
    (bicycle)
;

[Action]
    ([ActionSize] [ActionVerb])
    ([ActionVerb])
;

[ActionVerb]
    ([ActionVerbBig])
    ([ActionVerbSmall])
;

[ActionVerbSmall]
    ([ActionVerbSmallT])
    ([FootHook])
    ([Layback])
    ([Cross])
;

[ActionVerbSmallT]
    (bump)
    (out)
    (up)
    (left)
    (right)
    (fondle)
    (grab)
    (roll)
    (over)
    (diagonal)
    (slide)
    (grab)
    (drop)
    (go again)
    (go)
    (move)
;

[Cross]
    (cross over)
    (cross under)
    (crossover)
    (crossunder)
    (cross)
;


[ActionVerbBig]
    (throw)
    (dyno)
    (reach)
    (fall into)
    (huck)
    (deadpoint)
    (rock)
    (dead point)
;

[ActionSize]
    ([ActionSizeBig])
    ([ActionSizeSmall])
;

[ActionSizeBig]
    (big)
    (huge)
    (far)
;

[ActionSizeSmall]
    (small)
;
\end{verbatim}

\end{multicols}

\bibliography{tr}

\begin{thebibliography}{10}

\bibitem{Anderson2004}
L.~Anderson.
\newblock {\em The Art of Coursesetting}.
\newblock 2004.

\bibitem{RandomArt}
A.~Bauer.
\newblock Random art.
\newblock \url{http://www.random-art.org/}, December 2009.

\bibitem{Begleiter2004}
R.~Begleiter, R.~El-Yaniv, and G.~Yona.
\newblock On prediction using variable order {Markov} models.
\newblock {\em Journal of Artificial Intelligence Research}, 22:385--421, 2004.

\bibitem{Bell1989}
T.~Bell, I.~H. Witten, and J.~G. Cleary.
\newblock Modeling for text compression.
\newblock {\em ACM Comput. Surv.}, 21:557--591, December 1989.

\bibitem{Bethlehem2009}
J.~Bethlehem.
\newblock {\em Applied Survey Methods: A Statistical Perspective}.
\newblock Wiley, 2009.

\bibitem{bradley}
E.~Bradley, D.~Capps, J.~Luftig, and J.~M. Stuart.
\newblock Towards stylistic consonance in human movement synthesis.
\newblock {\em Open {AI} Journal}, 4:1--19, 2011.

\bibitem{bradley98a}
E.~Bradley and J.~Stuart.
\newblock Using chaos to generate variations on movement sequences.
\newblock {\em Chaos}, 8:800--807, 1998.

\bibitem{Briggs1992}
J.~Briggs.
\newblock {\em Fractals: The Patterns of Chaos}.
\newblock Touchstone, 1992.

\bibitem{Carifio2007}
J.~Carifio and R.~J. Perla.
\newblock Ten common misunderstandings, misconceptions, persistent myths and
  urban legends about {Likert} scales and {Likert} response formats and their
  antidotes.
\newblock {\em Journal of Social Sciences}, 3:106--116, 2007.

\bibitem{dabby}
D.~S. Dabby.
\newblock Musical variations from a chaotic mapping.
\newblock {\em Chaos: An Interdisciplinary Journal of Nonlinear Science},
  6(2):95--107, 1996.

\bibitem{Gleim2003}
J.~A. Gleim and R.~R. Gleim.
\newblock Calculating, interpreting, and reporting {Cronbach's} alpha
  reliability coefficient for {Likert}-type scales.
\newblock In {\em 2003 Midwest Research to Practice Conference in Adult,
  Continuing, and Community Education}, 2003.

\bibitem{Green2005}
J.~Green and J.~Brown.
\newblock {\em Principles of Social Research}.
\newblock Open University Press, 2005.

\bibitem{Hartley1995}
J.~Hartley.
\newblock Generative processes in algorithmic composition: Chaos and music.
\newblock {\em Leonardo}, 28:221--224, 1995.

\bibitem{Jamieson2004}
S.~Jamieson.
\newblock {Likert} scales: How to (ab)use them.
\newblock {\em Medical Education}, 38:1212--1218, 2004.

\bibitem{Openheim1966}
A.~Oppenheim.
\newblock {\em Questionnaire Design and Attitude Measurement}.
\newblock Basic Books Inc., 1966.

\bibitem{Palmer:2005fk}
M.~Palmer, D.~Gildea, and P.~Kingsbury.
\newblock The proposition bank: A corpus annotated with semantic roles.
\newblock {\em Computational Linguistics Journal}, 31(1), 2005.

\bibitem{ClimbingMagazine}
C.~Phillips.
\newblock Off the wall: Climbing with chaos.
\newblock {\em Climbing Magazine}, 282, October 2009.

\bibitem{Quaine1999}
F.~Quaine and L.~Martin.
\newblock A biomechanical study of equilibrium in sport rock climbing.
\newblock {\em Gait and Posture}, 10:233--239, 1999.

\bibitem{ORCA2003}
{Roper Research}.
\newblock Outdoor recreation in {America}.
\newblock Technical report, The Roper Center for Public Opinion Research,
  December 2003.

\bibitem{Shearer1992}
R.~R. Shearer.
\newblock Chaos theory and fractal geometry: Their potential impact on future
  art.
\newblock {\em Leonardo}, 25:143--152, 1992.

\bibitem{Smyth1998}
M.~M. Smyth and A.~Waller.
\newblock Movement imagery in rock climbing: Patterns of interference from
  visual, spatial and kinaesthetic secondary tasks.
\newblock {\em Applied Cognitive Psychology}, 12:145--157, 1998.

\bibitem{Streiner2003}
D.~L. Streiner.
\newblock Starting at the beginning: An introduction to coefficient alpha and
  internal consistency.
\newblock {\em Journal of Personality Assessment}, 80:99--103, 2003.

\bibitem{bradley98c}
J.~Stuart and E.~Bradley.
\newblock Learning the grammar of dance.
\newblock In {\em Proceedings of the International Conference on Machine
  Learning ({ICML})}, pages 547--555, 1998.

\bibitem{Ward1994}
W.~Ward.
\newblock Extracting information in spontaneous speech.
\newblock In {\em Third International Conference on Spoken Language Processing
  (ICSLP 94)}, 1994.

\bibitem{Watts2004}
P.~B. Watts.
\newblock Physiology of difficult rock climbing.
\newblock {\em European Journal of Applied Physiology}, 91:361--372, 2004.

\end{thebibliography}

\end{document}